\documentclass{article}

\usepackage[preprint]{corl_2024} 

\usepackage{graphicx}
\usepackage{amsmath}
\usepackage{amssymb}
\usepackage{mathtools}
\usepackage{amsthm}
\usepackage{multirow}
\usepackage{booktabs}
\usepackage{fontawesome5}
\usepackage{wrapfig}
\usepackage{lipsum}

\newtheorem{myrem}{Remark}

\usepackage[usenames,dvipsnames,table]{xcolor}
\usepackage{nicematrix}
\usepackage{float}

\title{
Robust RL with LLM-Driven Data Synthesis and Policy Adaptation for Autonomous Driving}

\author{
  \textbf{Sihao Wu$^{1\star}$, Jiaxu Liu$^{1\star}$, Xiangyu Yin$^{1}$, Guangliang Cheng$^{1}$, Xingyu Zhao$^{2}$,}\\
  \textbf{Meng Fang$^{1}$, Xinping Yi$^{3}$, Xiaowei Huang$^{1}$}\\
  $^{1}$University of Liverpool, $^{2}$WMG, University of Warwick, $^{3}$Southeast University\\
  \texttt{\{firstname.lastname\}}@liverpool.ac.uk, xingyu.zhao@warwick.ac.uk, xyi@seu.edu.cn \\
}

\begin{document}
\maketitle


\begin{abstract}
The integration of Large Language Models (LLMs) into autonomous driving systems demonstrates strong common sense and reasoning abilities, effectively addressing the pitfalls of purely data-driven methods. Current LLM-based agents require lengthy inference times and face challenges in interacting with real-time autonomous driving environments. A key open question is whether we can effectively leverage the knowledge from LLMs to train an efficient and robust Reinforcement Learning (RL) agent. This paper introduces RAPID, a novel \underline{\textbf{R}}obust \underline{\textbf{A}}daptive \underline{\textbf{P}}olicy \underline{\textbf{I}}nfusion and \underline{\textbf{D}}istillation framework, which trains specialized mix-of-policy RL agents using data synthesized by an LLM-based driving agent and online adaptation. RAPID features three key designs: 1) utilization of offline data collected from an LLM agent to distil expert knowledge into RL policies for faster real-time inference; 2) introduction of robust distillation in RL to inherit both performance and robustness from LLM-based teacher; and 3) employment of a mix-of-policy approach for joint decision decoding with a policy adapter. Through fine-tuning via online environment interaction, RAPID reduces the forgetting of LLM knowledge while maintaining adaptability to different tasks. Extensive experiments demonstrate RAPID's capability to effectively integrate LLM knowledge into scaled-down RL policies in an efficient, adaptable, and robust way. Code and checkpoints will be made publicly available upon acceptance.

\end{abstract}


\section{Introduction}

The integration of Large Language Models (LLMs) with emergent capabilities into autonomous driving presents an innovative approach \cite{tian2024drivevlm,jin2023surrealdriver,jin2023adapt}. 
Previous work suggests that LLM can significantly enhance the common sense and reasoning abilities of autonomous vehicles, effectively addressing several pitfalls of purely data-driven methods \cite{fu2023drive, wen2023dilu,chen2023driving}. However, LLMs face several challenges, primarily in generating effective end-to-end instructions in real-time and dynamic driving scenarios. This limitation stems from two primary factors: the extended inference time required by LLM-based agents \cite{wen2023dilu} and the difficulty these agents face in continuous data collection and learning \cite{carta2023grounding}, which renders them unsuitable for real-time decision-making in dynamic driving environments. 
Furthermore, faster and smaller models, which are often preferred for real-time applications, have a higher risk of being vulnerable to adversarial attacks compared to larger models \cite{Huang_2023_CVPR, zi2021revisiting, goldblum2020adversarially, zhu2021reliable}. These challenges drive us to tackle the following questions:


\begin{quote}
\centering
How to develop an \textbf{efficient}, \textbf{adaptable}, and \textbf{robust} agent that can leverage the capabilities of the LLM-based agent for autonomous driving? 
\end{quote}

One potential solution is to use the LLM as a teacher policy to instruct the learning of a lighter, specialized student RL policy through knowledge distillation~\cite{shi2023unleashing,zhou2023large}. This allows the student model to inherit the reasoning abilities of the LLM while being lightweight enough for real-time inference. Another approach is to employ LLMs to generate high-level plans or instructions, which are then executed by a separate controller~\cite{huang2022language}. This decouples the reasoning and execution processes, allowing for faster reaction times. Furthermore, techniques such as Instruction Tuning (IT) \cite{wei2022finetuned} and In-Context Learning (ICL) \cite{brown2020language} can adapt the LLM to new tasks without extensive fine-tuning. However, these approaches still have limitations. Knowledge distillation may result in information forgetting or loss of generalization ability. Generating high-level plans relies on the strong assumption that the LLM can provide complete and accurate instructions. IT and ICL are sensitive to the choice of prompts and demonstrations, which require careful design for each task \cite{wang2022zemi,gu2022learning}.

To tackle the above challenges, we propose RAPID, a \underline{\textbf{R}}obust \underline{\textbf{A}}daptive \underline{\textbf{P}}olicy \underline{\textbf{I}}nfusion and \underline{\textbf{D}}istillation framework that incorporates LLM knowledge into offline RL for autonomous driving, to leverage the common sense and robust ability of LLM and solve challenging scenarios such as unseen corner cases. Our method encompasses several designs:
\textit{(1)} We utilize offline data collected from LLMs-based agents to facilitate the distillation of expertise information into faster policies for real-time inference.
\textit{(2)} We introduce robust distillation in offline RL to inherit not only performance but also the robustness from LLM teachers. 
\textit{(3)} We introduce mix-of-policy for joint action decoding with policy adapter. Through fine-tuning via online environment interaction, we prevent the forgetting of LLM knowledge while keeping the
adaptability to various RL environments. 

To the best of our knowledge, this work pioneers the distillation of knowledge from LLM-based agents into RL policies through offline training combined with online adaptation in the context of autonomous driving. Through extensive experiments, we demonstrate RAPID's capability to effectively integrate LLM knowledge into scaled-down RL policies in a transferable, robust, and efficient way. 

\section{Preliminaries}

\textbf{Notation.} We view a sequential decision-making problem, formalized as a Markov Decision Process (MDP), denoted by $\langle \mathcal{S}, \mathcal{A}, \mathcal{T}, \mathcal{R}, \gamma \rangle$, where $\mathcal{S}$ and $\mathcal{A}$ represent the state and action spaces, respectively. The transition probability function is denoted by $\mathcal{T}: \mathcal{S} \times \mathcal{A} \rightarrow \mathcal{P}(S)$, and the reward function is denoted by $\mathcal{R}: \mathcal{S} \times \mathcal{A} \times \mathcal{S} \rightarrow \mathbb{R}$. Moreover, $\gamma$ denotes the discount factor. The main objective is to acquire an optimal policy, denoted as $\pi: \mathcal{S} \rightarrow \mathcal{A}$, that maximizes the expected cumulative return over time, $\max_{\pi} \mathbb{E}[\sum_{t} \gamma^{t} r_{t}]$. The policy parameter $\theta$, denoted in $\pi$, is crucial. A typical gradient-based RL algorithm minimizes a surrogate loss, $J (\theta)$, employing gradient descent concerning $\theta$. This loss is estimated using sampled trajectories, wherein each trajectory comprises a sequence of state-action-reward tuples.

\textbf{Offline RL.} The aim is to learn effective policies from pre-collected datasets, eliminating the need for further interaction. Given a dataset $\mathcal{D} = \{(\mathbf{s}, \mathbf{a}, r, \mathbf{s^{\prime}})\}$ containing trajectories collected under an unknown behavior policy
$\pi_{B}$,  the iterative Q update step with learned policy $\pi$ is expressed as
\begin{align}
    &\hat{Q}^{k+1} \gets \arg\min_{Q} J(Q, \pi, \mathcal{D}), \quad \text{(Policy Evaluation)} \label{eq:policy_evaluation} \\
    &\text{where } J(Q, \pi, \mathcal{D}) := \mathbb{E}_{\mathbf{s},\mathbf{a},\mathbf{s}' \sim \mathcal{D}} \left[ \big( \big( r(\mathbf{s},\mathbf{a})  + \gamma \mathbb{E}_{\mathbf{a}' \sim \hat{\pi}(\mathbf{a}'|\mathbf{s}')} [\hat{Q}^k (\mathbf{s}', \mathbf{a}')] \big) - Q(\mathbf{s},\mathbf{a}) \big)^2 \right].
\end{align}
With the updated Q-function, the policy is improved by
\begin{align}
    \hat{\pi}^{k+1} \gets \arg\max_{\pi} \mathbb{E}_{\mathbf{s}\sim\mathcal{D}, \mathbf{a}\sim \pi^k(\mathbf{a}|\mathbf{s})} \left[
    \hat{Q}^{k+1}(\mathbf{s}, \mathbf{a})
    \right]. \label{eq:policy-improvement} \quad \text{(Policy Improvement)}
\end{align}
\textbf{Conservative Q-Learning.} Offline RL algorithms following this fundamental approach are often challenged by the issue of action distribution shift \cite{kumar2019stabilizing}. Therefore, \cite{kumar2020conservative} proposed conservative Q-learning, where the Q values are penalized by Out-Of-Distribution (OOD) actions
\begin{align}
    \hat{Q}^{k+1} \gets \arg\min_{Q} J(Q, \pi, \mathcal{D}) + \alpha \left( \mathbb{E}_{\mathbf{s} \sim \mathcal{D}, \mathbf{a} \sim \mu(\cdot | s)} [ Q (\mathbf{s}, \mathbf{a}) ]  - \mathbb{E}_{\mathbf{s},\mathbf{a} \sim \mathcal{D}} [Q (\mathbf{s}, \mathbf{a})]\right) \label{eq:conservative_offline},
\end{align}
where $\mu$ is an approximation of the policy that maximizes the current Q-function. While \cite{schweighofer2022dataset} found that the effectiveness of Offline RL algorithms is significantly influenced by the characteristics of the dataset, which also motivated us to explore the influence of LLM-generated dataset.










\section{RAPID: Robust Distillation and Adaptive Policy Infusion}
\vspace{-5pt}

\subsection{Offline Dataset Collection}

\begin{figure*}[t]
\centering
\includegraphics[width=1\linewidth]{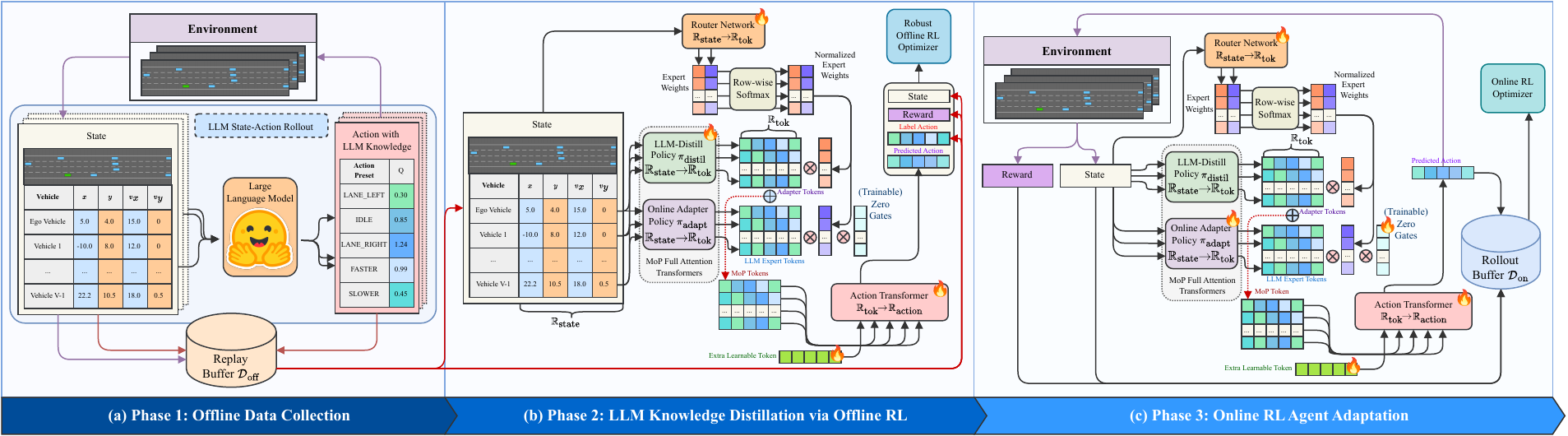}
\caption{Our proposed RAPID framework. Only modules tagged by \textcolor{RedOrange}{\faFire*} are trained. \textit{(a) Phase 1}: Collect state-action rollouts from an environment and store them in a replay buffer. \textit{(b) Phase 2}: Distill LLM knowledge into an offline policy using the collected data, the adapter policy is frozen and its output tokens are masked by zero gates. \textit{(c) Phase 3}: Adapt the pre-trained model online by interacting with the environment, the LLM-distilled policy is frozen, and the zero gate is trained for progressive adaptation.}
\label{figure:training-pipeline}
\end{figure*}

As shown in Fig.~\ref{figure:training-pipeline} (a), we conducted a closed-loop driving experiment on HighwayEnv \cite{highway-env} using \texttt{GPT-3.5} \cite{chatgpt} to collect the offline dataset. The vehicle density and number of lanes in HighwayEnv can be adjusted, and we choose \textsc{lane-3-density-2} as the base environment. As a text-only LLM, \texttt{GPT-3.5} cannot directly interact with the HighwayEnv simulator. 
To facilitate its observation and decision-making processes, the experiment incorporated perception tools and agent prompts, enabling \texttt{GPT-3.5} to effectively engage with the simulated environment. The prompts have the following stages: \textit{(1)} Prefix prompt: The LLM obtains the current driving scene and historical information. \textit{(2)} Reasoning: By employing the ReAct framework \cite{yao2022react}, the LLM-based agent reasons about the appropriate driving actions based on the scene.  \textit{(3)} Output decision: The LLM outputs its decision on which meta-action to take. The agent has access to $5$ meta-actions: \texttt{lane\_left}, \texttt{lane\_right}, \texttt{faster}, \texttt{slower}, and \texttt{idle}. More details about the prompt setup are instructed in Appendix~\ref{app:prompt_setup}. Through an iterative closed-loop process described above, we collect the dataset $\mathcal{D}_{\mathrm{LLM}} = \{(\mathbf{s}, \mathbf{a}, \mathbf{r}, \mathbf{s^{\prime}}) | \mathbf{a} \sim \pi_{\mathrm{LLM}}(\mathbf{a}|\mathbf{s})\}$, where the $\pi_{\mathrm{LLM}}$ is the LLM agent.






\begin{wrapfigure}{r}{0.4\textwidth}
\centering
\vspace{-15pt}
\includegraphics[width=\linewidth]{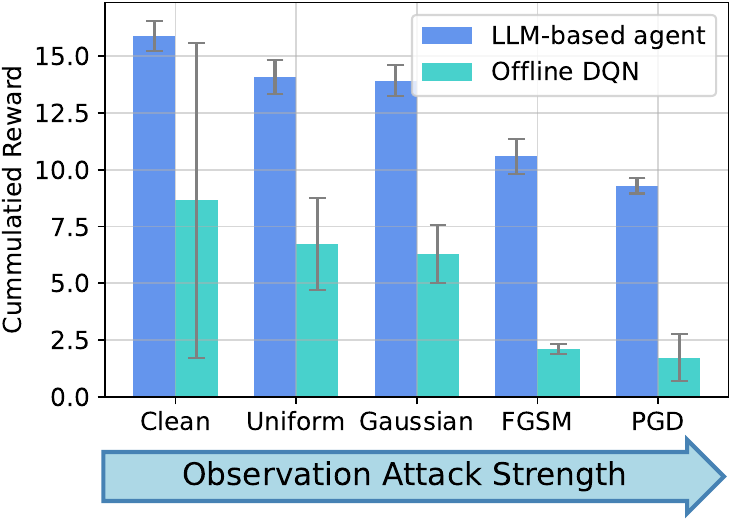}
\vfill
\vspace{-10pt}
\caption{Cumulative reward on \textsc{lane-3-density-2} with progressively increasing observation ($\mathbf{s}$) attack strength. The LLM-based agent $\pi_{\mathrm{LLM}}$ exhibits improved robustness compared with vanilla Offline DQN.}
\label{figure:llm_attack_demo}
\end{wrapfigure}
 
\subsection{Robustness Regularized Distillation}
Recall the offline RL objective in Eq.~(\ref{eq:policy-improvement}-\ref{eq:conservative_offline}). Let the LLM-distil policy be $\pi_{\mathrm{distil}}$, with the collected dataset $\mathcal{D}_\mathrm{LLM}$, the offline training is to optimize $J(Q, \pi_{\mathrm{distill}}, \mathcal{D}_{\mathrm{LLM}})$ for an improved $Q$ function, then update the policy \textit{w.r.t} $Q$. 
Empirically as shown in Fig.~\ref{figure:llm_attack_demo}, the LLM-based agent $\pi_{\mathrm{LLM}}$ is more robust against malicious state injection under the autonomous driving setting. However, a distilled offline policy is not as robust compared to LLMs, demonstrated by \cite{yang2022rorl}, where the $Q$ value can change drastically over neighbouring states, leading to an unstable policy. Therefore, vanilla offline RL algorithms cannot robustly distil information to the LLM-distil agent.   
Inspired by ~\cite{goldblum2020adversarially, zhu2021reliable}, we formulate a novel training objective by introducing a discrepancy term into Eq.~(\ref{eq:conservative_offline}), allowing the distillation of adversarial knowledge to the offline agent.
\begin{align}
    &J_\mathrm{robust}(Q, \pi, \mathcal{D}) :=  J(Q, \pi, \mathcal{D}) + \alpha \left(
    \mathbb{E}_{\mathbf{s} \sim \mathcal{D}, \mathbf{a} \sim \mu(\cdot | \mathbf{s})} \left[ Q (\mathbf{s}, \mathbf{a}) \right]  - 
    \mathbb{E}_{\mathbf{s},\mathbf{a} \sim \mathcal{D}} \left[Q (\mathbf{s}, \mathbf{a})\right]
    \right)\label{eq:robust_distill}\\
    & + \textcolor{blue}{
    \beta \left(
    \mathbb{E}_{\mathbf{s},\mathbf{a} \sim \mathcal{D}}\left[
    \log\left(\frac{
    \sigma\left(Q\left(\tilde{\mathbf{s}}, \mathbf{a}\right)\right)
    }{ 
    \mathrm{onehot}(\mathbf{a})
    }\right)
    \right]
    \right), \text{ where } \tilde{\mathbf{s}} = \mathop{\arg\max}_{{\|\tilde{\mathbf{s}}-\mathbf{s}\|_2\leq\epsilon}} 
    \mathbb{E}_{\mathbf{s},\mathbf{a} \sim \mathcal{D}}\left[
    \log\left(\frac{
    \sigma\left(Q\left(\tilde{\mathbf{s}}, \mathbf{a}\right)\right)
    }{ 
    \mathrm{onehot}(\mathbf{a})
    }\right)
    \right]
    }
    . \nonumber
\end{align}
In Eq.~(\ref{eq:robust_distill}), $\sigma(\cdot)$ denote the $\mathrm{softmax}(\cdot)$ function, characterizing the probability that the Q network will choose each action. $\mathrm{onehot}(\cdot)$ converts the selected action from the offline dataset to a one-hot vector, characterizing the definite events. Therefore, $
    \mathbb{E}_{\mathbf{s},\mathbf{a} \sim \mathcal{D}}\left[
    \log\left(\frac{
    \sigma\left(Q\left(\tilde{\mathbf{s}}, \mathbf{a}\right)\right)
    }{ 
    \mathrm{onehot}(\mathbf{a})
    }\right)
\right]$ can be viewed as the KL divergence between the \textit{$Q$ network output distribution} and the \textit{categorical distribution of one-hot action}. Essentially, the $\arg\max$ constraint identifies the adversarial state that yields the worst $Q$ performance, while the objective $J_\mathrm{robust}$ seeks to find the optimal $Q$ function to neutralize the adversarial attack. This process forms an adversarial training procedure, where $\alpha$ and $\beta$ are hyperparameters that balance the conservative and robustness terms, respectively.

\subsection{LLM Knowledge Infused Robust Policy with Environment Adaptation}\label{RAPID}
In autonomous driving, the RL policy typically has the overview of the ego car, as well as surrounding cars' information. The ego action is predicted by considering the motion of all captured cars. Assuming $V$ vehicles captured, $F$ vehicle features
%
and $A$ action features\footnote{features: \url{highway-env.farama.org/observations}, actions: \url{highway-env.farama.org/actions}}
, the state can be encapsulated by a matrix in $\mathbb{R}^{V\times F}$, and the action is simply a vector in $\mathbb{R}^{A}$. A typical way to obtain actions from observations is to concat all $V$ features in a row, then feed into a  $\mathrm{MLP}:\mathbb{R}^{V*F} \to \mathbb{R}^{A}$. However, when the vanilla RL agent is expanded to multi-policy for joint decision, \textit{i.e.,} the action is modelled via interdisciplinary collaboration (\textit{e.g. LLM-distilled knowledge \& online environment}). In such a case, simple MLPs encode vehicles as one unified embedding, which fails to model the explicit decision process for each other vehicle, thus lacking in explainability. Below, we discuss how to incorporate different sources of knowledge for joint decision prediction.

\subsubsection{Mixture-of-Policies via Vehicle Features Tokenization}

Let the observation $\mathbf{s}\in \mathbb{R}^{V\times F}$, assume we have $N$ policies for joint decision. Our approach is to revise the state $\mathbf{s}$ as a sequence of tokens, where the sequence is of length $V$ and each token is of dimension $F$. Borrowing ideas from language models \cite{clark2022unified,hazimeh2021dselect,zhou2022mixture}, we implement \textit{mixture-of-policies} (MoP) for joint-decision. As detailed below, we illustrate the process to obtain action $a$ from state $s$. Assume the $N$ policies $\pi_{1...N}: \mathbb{R}^{V\times F}\to \mathbb{R}^{V\times D}$ where $D$ is the latent token dim. The state is first fed into a router network $G: \mathbb{R}^{V\times F}\to \mathbb{R}^{V\times N}$ to get the policy weights $G(\mathbf{s}) \in \mathbb{R}^{V\times N}$. Then a $\mathrm{topK}$ and $\mathrm{Softmax}$ is applied column-wise to select the most influential $K$ policies and get the normalization $\omega = \mathrm{softmax}(\mathrm{topK}(G(\mathbf{s}))) \in \mathbb{R}^{V\times N}$. Next, we calculate output sequences of all policies: $\mathbf{t} = \{\pi_i(\mathbf{s})\}_{i=1}^N \in \mathbb{R}^{N\times V \times D}$. The mixed policy is then the weighted mean of sequences, expressed as $\Tilde{\mathbf{a}} = \sum_{i=1}^N (\omega^\top)_i \cdot \mathbf{t}_i\in \mathbb{R}^{V \times D}$. Finally the action is obtained via action decoder $\mathrm{dec}: \mathbb{R}^{V \times D}\to \mathbb{R}^{A}$. In one equation, the joint policy $\mathbf{s}\to \mathbf{a}$ is formulated by
\begin{align}
    \mathbf{a} = \Pi_\mathrm{MoP}(\mathbf{s}; \theta_\mathrm{d}, \theta_\mathrm{r}, \theta_\mathrm{p}):= \mathrm{dec}_{\theta_\mathrm{d}}\left(
        \sum_{i=1}^N \left[ \left(\left[\mathrm{softmax}\left(\mathrm{topK}\left(G_{\theta_\mathrm{r}}\left(\mathbf{s} \right)\right)\right)\right]^\top\right)_i \cdot \pi_i\left(\mathbf{s}; \theta_\mathrm{p}^{(i)}\right) \right]
    \right) \label{eq:mix_of_policies}
\end{align}
where $\theta_\mathrm{d}, \theta_\mathrm{r}, \theta_\mathrm{p}$ are respectively the parameters of action decoder, router, and policies. $\Pi_\mathrm{MoP}$ represents the mixed policy for the joint decision from both distilled policy and online adaptation policy. In practice, we design the policies as full-attention transformers, and decoder $\mathrm{dec}_{\theta_d}$ as a ViT-wise transformer, taking $1+V$ tokens as input sequence where the first token is an extra learnable token. The extra token's embedding is decoded as the predicted action. For detailed implementation of MoP policies and the action decoder, please refer to Appendix~\ref{app:nets_implementation}.
\begin{myrem}
    One simple alternative to joint action prediction is to mix actions instead of policies. Regarding $N$ policies as $\mathrm{mlp}_{1...N}: \mathbb{R}^{V*F} \to \mathbb{R}^{A}$, the final action $a\in \mathbb{R}^{A}$ is then the merge of their respective decisions, \textit{i.e.} $\mathbf{a} = \mathrm{merge}(\mathrm{mlp}_{1...N}(\mathrm{flat}(\mathbf{s})); w_{1...N})$ where $\mathrm{flat}:\mathbb{R}^{V\times F}\to \mathbb{R}^{V*F}$. However, this approach presents several problems: \textbf{Q1}: Weights $w_{1...N}$, despite learnable, are fixed vehicle-wise. This means all vehicles in policy $1$ share the same $w_1$ for action prediction, same as for other policies. \textbf{Q2}: Weights are independent of observations, which are not generalizable to constantly varying environments. \textbf{Q3}: The $\mathrm{merge}$ is not a sparse selection, meaning all policy candidates are selected for action decision, resulting in loss of computational efficiency when $N$ increases. Using Eq.~(\ref{eq:mix_of_policies}), as $\omega^\top\in \mathbb{R}^{N\times V}$, for each policy, we have $V$ different weights for each vehicle, this resolves \textbf{Q1}. Meanwhile, as $\omega$ is generated from routed state $G(\mathbf{s})$, which is adaptive to varying states, this resolves \textbf{Q2}. \textbf{Q3} is also resolved through employing $\mathrm{topK}$, when $K< N$, the selection on policies will be sparse.
\end{myrem}

\subsubsection{Online Adaptation to Offline Policy with Mixture-of-Policies}
Our employment of Eq.~(\ref{eq:mix_of_policies}) is illustrated in Fig.~\ref{figure:training-pipeline}, a special case where $N=K=2$. The policies come from two sources, respectively, the distilled language model knowledge (\textit{LLM-distill Policy}, $\pi_1=\pi_\mathrm{distil}$) and the online environment interaction (\textit{Online Adapter Policy}, $\pi_2=\pi_\mathrm{adapt}$). The robust distillation principal (Eq.~(\ref{eq:robust_distill})) is integrated into the offline distillation phase (Fig.~\ref{figure:training-pipeline}.(b)). Respectively, the objective for Q-network \& joint policy update is expressed as
\begin{align}
    & \hat{Q}^{k+1} \gets \arg\min_{Q} J_\mathrm{robust}\left( Q, \Pi_\mathrm{MoP}(\cdot|\cdot; \theta_\mathrm{d}, \theta_\mathrm{r}, \theta_\mathrm{p}), \mathcal{D}_\mathrm{off} \right),  \text{ (Robust Offline Q-Learning)}\\
    & \hat{\Pi}^{k+1}_\mathrm{MoP} \gets \mathrm{arg}\max_{\theta_\mathrm{d}, \theta_\mathrm{r}, \theta_\mathrm{p}^{(1)}} \mathbb{E}_{\mathbf{s}\sim \mathcal{D}_\mathrm{off}, \mathbf{a}\sim \Pi^k_\mathrm{MoP}(\cdot|\mathbf{s}; \theta_\mathrm{d}, \theta_\mathrm{r}, \theta_\mathrm{p})} \left[
        \hat{Q}^{k+1} (\mathbf{s}, \mathbf{a})
    \right]. \label{eq:offline_objective}  \text{ (LLM Policy Improvement)}
\end{align}
The parameters $\theta_p^{(2)}$ of $\pi_2$ ($\pi_\mathrm{adapt}$) during policy improvement (Eq.~{(\ref{eq:offline_objective})}) are frozen since LLM knowledge should only be distilled into $\pi_1$ ($\pi_\mathrm{distil}$). With an arbitrary RL algorithm, the online adaptation phase objective can be expressed by
\begin{align}
    \hat{\Pi}^\star_\mathrm{MoP} = \mathrm{arg}\max_{\theta_\mathrm{d}, \theta_\mathrm{r}, \theta_\mathrm{p}^{(2)}} \mathbb{E}_{\mathbf{s}\sim \mathcal{D}_\mathrm{on}, \mathbf{a}\sim \Pi_\mathrm{MoP}(\cdot|\mathbf{s}; \theta_\mathrm{d}, \theta_\mathrm{r}, \theta_\mathrm{p})} \left[
        Q(\mathbf{s}, \mathbf{a}) 
    \right]. \label{eq:online_objective} \quad\text{(Adapter Policy Optimization)}
\end{align}
In Eq.~{(\ref{eq:online_objective})}, the learned policy interacts with the environment rolling out the $(\mathbf{s}, \mathbf{a}, r)$ tuple. With the frozen distilled knowledge of LLM in $\pi_\mathrm{distil}$, parameterized by $\theta_p^{(1)}$, we fine-tune $\pi_\mathrm{adapt}$, parameterized by $\theta_p^{(2)}$, to adapt the MoP policy $\Pi_\mathrm{MoP}$ with LLM prior to the actual RL environment.

\subsubsection{Zero Gating Adapter Policy}
To eliminate the influence of $\pi_\mathrm{adapt}$ on $\pi_\mathrm{distil}$ during the offline phase, we adopt the concept of zero gating \cite{zhang2023llamaadapter} for initializing the policy. Specifically, the router network's corresponding expert weights for $\pi_\mathrm{adapt}$ are masked with a trainable zero vector. As a result, the MoP token in phase 2 only considers the impact of $\pi_\mathrm{distil}(\mathbf{s})$. During phase 3, we enable the training of zero gates, allowing adaptation tokens to progressively inject newly acquired online signals into the MoP policy $\Pi_\mathrm{MoP}$.

\section{Experiments}



\subsection{Experiment Setting}\label{section:experiment_setting}
We conduct experiments using the HighwayEnv simulation environment \cite{highway-env}. We involve three driving scenarios with increasing levels of complexity: \textsc{lane-3-density-2}, \textsc{lane-4-density-2.5}, and \textsc{lane-5-density-3}. Detailed task descriptions are delegated to Appendix~\ref{app:env_details}. This paper constructs three types of datasets, namely the \textit{random} $\mathcal{D}_\mathrm{rand}$, the \textit{LLM-collected} $\mathcal{D}_\mathrm{LLM}$, and the \textit{combined} dataset $\mathcal{D}_\mathrm{off}^\star$. The construction details and optimal ratio for $\mathcal{D}_\mathrm{off}^\star$ are discussed in Sec.~\ref{section:effectiveness_of_LLM_generated_dataset}.

We compare the RAPID performance under the offline phase with several state-of-the-art RL methods, DQN \cite{mnih2015human}, DDQN \cite{van2016deep}, and CQL \cite{kumar2020conservative} respectively. In the online phase, we employ the DQN algorithm as the adaptation method under our RAPID framework. To validate the efficacy of $J_{\mathrm{robust}}$ term in Eq.~(\ref{eq:robust_distill}), we utilize four various attack methods: Uniform, Gaussian, FGSM, and PGD, to evaluate the robustness of the distilled policy. All baselines are implemented with d3rlpy library \cite{seno2022d3rlpy}. The hyperparameters setting refers to Appendix.~\ref{app:hyperparameters}. Each method is trained for a total of 10K training iterations, with evaluations performed per 1K iterations. We employ the same reward function as defined in highway-env-rewards\footnote{rewards: \url{highway-env.farama.org/rewards}}.


\subsection{Fusing LLM Generated Dataset (Phase 1)}
\label{section:effectiveness_of_LLM_generated_dataset}

\begin{wrapfigure}{r}{0.5\textwidth}
\centering
\vspace{-10pt}
\includegraphics[width=\linewidth]{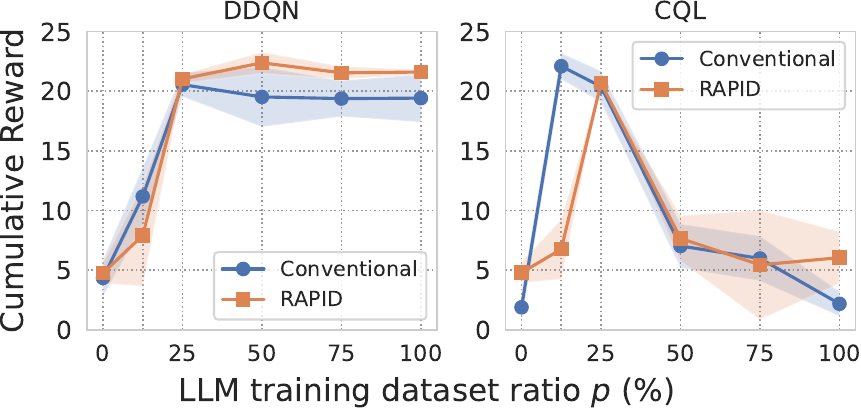}
\vfill
\caption{Cumulative reward over different training ratios under offline training framework in \textsc{highway-fast} environment. The result is averaged over $5$ random trails.}
\vspace{-10pt}
\label{figure:plot_ratio}
\end{wrapfigure}

To validate the efficacy of the LLM-collected dataset $\mathcal{D}_\mathrm{LLM}$, we build the offline dataset by combining it with a random dataset $\mathcal{D}_\mathrm{rand}$. We evaluate the cumulative rewards on $\mathcal{D}_\mathrm{off} = \mathrm{sample}(\{\mathcal{D}_\mathrm{LLM}, \mathcal{D}_\mathrm{rand}\}; \{p, 1-p\})$ with two offline algorithms: DDQN \cite{van2016deep} and CQL \cite{kumar2020conservative}, to find the best ratio $p$. 
To clarify, $\mathcal{D}_\mathrm{rand}$ is sampled using a random behavioural policy, serving as a baseline for data collection. $\mathcal{D}_\mathrm{LLM}$ is sampled via a pre-trained LLM $\pi_\mathrm{LLM}$. We gather 3K transitions using $\pi_\mathrm{LLM}$ for each trail. Therefore, both $\mathcal{D}_\mathrm{random}$ and $\mathcal{D}_\mathrm{LLM}$ dataset contains 15K transitions.

According to Fig.~\ref{figure:plot_ratio}, the pure $\mathcal{D}_\mathrm{rand}$ with $p=0\%$ cannot support a well-trained offline policy. Utilizing only the $\mathcal{D}_\mathrm{LLM}$ can also harm the performance. However, augmenting offline datasets with partial LLM-generated data can significantly boost policy cumulative reward.  With such evidence, we choose $p=25\%$ in between $12.5\%\sim50\%$ as a sweet spot and define it as our final $\mathcal{D}_\mathrm{off}^\star$.

\subsection{Offline LLM Knowledge Distillation (Phase 2)}
\label{section:distillation_benchmark}
\vspace{-5pt}

\begin{table}[H]
\centering
\caption{The cumulative rewards in \textbf{Phase 2}, averaged over $3$ seeds. RAPID (with\&without $J_{\mathrm{robust}}$) only enable the $\pi_\mathrm{distil}$ part (as depicted in Fig.~\ref{figure:training-pipeline}) in offline training. RAPID (w/o $J_{\mathrm{robust}}$) distils the LLM knowledge with Eq.~(\ref{eq:conservative_offline}) while RAPID uses robust distillation Eq.~(\ref{eq:robust_distill}). \colorbox{Cyan!30}{\strut Blue} / \colorbox{BurntOrange!30}{\strut Orange} indicate the first/second highest return for each buffer. \textbf{Bold} denotes the best within each environment. 
}
\vspace{-5pt}
\resizebox{\linewidth}{!}
{\begin{NiceTabular}{@{}c|c|ccc|cc@{}}[colortbl-like]
\toprule
Environments                                 & \begin{tabular}[c]{@{}c@{}}Dataset\end{tabular} & CQL                & DQN                & DDQN               & RAPID \tiny{(w/o $J_{\mathrm{robust}}$)} & RAPID                                    \\ \midrule
\multirow{3}{*}{\textsc{lane-3-density-2}}   & $\mathcal{D}_\mathrm{rand}$                             & 3.01$_{\pm 1.49}$  & \cellcolor{BurntOrange!30} 4.40$_{\pm 1.57}$  & \cellcolor{Cyan!30} 4.86$_{\pm 1.58}$  & 3.47$_{\pm 1.45}$                        & 3.66$_{\pm 1.58}$                        \\
                                             & $\mathcal{D}_\mathrm{LLM}$                              & 7.96$_{\pm 1.63}$  & 9.14$_{\pm 8.80}$  & 10.83$_{\pm 4.94}$ & \cellcolor{BurntOrange!30} 12.22$_{\pm 7.03}$                       & \cellcolor{Cyan!30} 14.18$_{\pm 2.43}$                       \\
                                             & $\mathcal{D}_\mathrm{off}^\star$                        & \cellcolor{BurntOrange!30} 13.06$_{\pm 1.11}$ & 8.66$_{\pm 6.93}$  & 12.22$_{\pm 7.03}$ & 12.72$_{\pm 0.59}$                       & \cellcolor{Cyan!30} \textbf{15.33$_{\pm 5.30}$} \\ \midrule
\multirow{3}{*}{\textsc{lane-4-density-2.5}} & $\mathcal{D}_\mathrm{rand}$                             & \cellcolor{Cyan!30} 3.84$_{\pm 2.59}$  & 2.41$_{\pm 0.92}$  & 1.96$_{\pm 0.52}$  & 2.18$_{\pm 0.59}$                        & \cellcolor{BurntOrange!30} 3.60$_{\pm 2.67}$                        \\
                                             & $\mathcal{D}_\mathrm{LLM}$                              & 2.56$_{\pm 0.90}$  & 3.99 $_{\pm 1.18}$ & \cellcolor{Cyan!30} 5.82$_{\pm 6.16}$  & 3.27$_{\pm 2.05}$                        & \cellcolor{BurntOrange!30} 4.60$_{\pm 2.69}$                        \\
                                             & $\mathcal{D}_\mathrm{off}^\star$                        & 7.36$_{\pm 0.60}$  & 7.14$_{\pm 6.12}$  & 7.61$_{\pm 6.76}$  & \cellcolor{BurntOrange!30} 10.19$_{\pm 1.30}$                       & \cellcolor{Cyan!30} \textbf{10.29$_{\pm 0.20}$} \\ \midrule
\multirow{3}{*}{\textsc{lane-5-density-3}}   & $\mathcal{D}_\mathrm{rand}$                             & 1.53$_{\pm 0.29}$  & \cellcolor{Cyan!30} 5.59$_{\pm 2.84}$  & \cellcolor{BurntOrange!30} 3.72$_{\pm 2.85}$  & 3.13$_{\pm 0.36}$                        & 2.16$_{\pm 0.36}$                        \\
                                             & $\mathcal{D}_\mathrm{LLM}$                              & 2.23$_{\pm 0.29}$  & \cellcolor{Cyan!30} 5.54$_{\pm 7.98}$  & 4.76$_{\pm 1.89}$  & \cellcolor{BurntOrange!30} 5.26$_{\pm 4.61}$                        & 1.45$_{\pm 2.11}$                        \\
                                             & $\mathcal{D}_\mathrm{off}^\star$                        & 5.58$_{\pm 3.93}$  & \cellcolor{BurntOrange!30} 5.99$_{\pm 4.10}$  & 5.05$_{\pm 2.06}$  & 5.38$_{\pm 0.63}$                        & \cellcolor{Cyan!30} \textbf{6.14$_{\pm 2.85}$}    \\ \bottomrule
\end{NiceTabular}}
\label{table:distillation_benchmark}
\end{table}

Phase 2 only uses $\pi_\mathrm{distil}$ for offline training as described in Sec.~\ref{RAPID}. We evaluate RAPID on three environments, respectively, \textsc{lane-3-density-2}, \textsc{lane-4-density-2.5} and \textsc{lane-5-density-3} with three different types of datasets: $\mathcal{D}_\mathrm{rand}$, $\mathcal{D}_\mathrm{LLM}$ and $\mathcal{D}_\mathrm{off}^\star$ as described in Sec.~\ref{section:effectiveness_of_LLM_generated_dataset}. 
In particular, we collect $\mathcal{D}_\mathrm{off}^\star$ with \textsc{lane-3-density-2} only, and apply this dataset to trails \textsc{lane-3-density-2}-$\mathcal{D}_\mathrm{off}^\star$, \textsc{lane-4-density-2.5}-$\mathcal{D}_\mathrm{off}^\star$, and \textsc{lane-5-density-3}-$\mathcal{D}_\mathrm{off}^\star$. 

We present the offline results in Tab.~\ref{table:distillation_benchmark} and observe the following: \textit{(1)} With $\mathcal{D}_\mathrm{LLM}$, policies exhibit better offline performance than $\mathcal{D}_\mathrm{rand}$ in general, while $\mathcal{D}_\mathrm{off}^\star$, as a mixture of above, improve upon both randomly- and LLM-generated datasets. This again confirms our conclusion in Sec.~\ref{section:effectiveness_of_LLM_generated_dataset}. \textit{(2)} Our approach, RAPID, consistently outperforms conventional methods, with RAPID using the mixed $\mathcal{D}_\mathrm{off}^\star$ achieving the highest rewards overall. \textit{(3)} $J_{\mathrm{robust}}$ does not impact the clean performance under the offline training phase. \textit{(4)} $\mathcal{D}_\mathrm{off}^\star$ collected from \textsc{lane-3-density-2} achieves better performance across all tasks, indicating that the LLM-generated dataset contains general knowledge applicable to different tasks with the same state and action spaces.

\subsection{Online Adaptation Performance (Phase 3)}
\label{section:online_adaptation_performance}

\begin{figure}[t]
\centering
\includegraphics[width=1\linewidth]{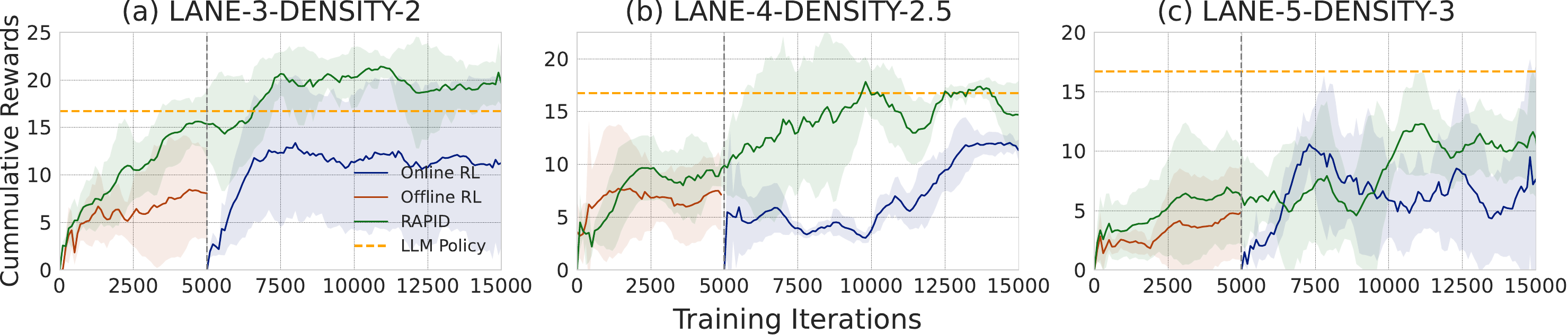}
\vfill
\caption{Performance of online adaptation (\textbf{Phase 3}) across three environments. Before the 5K-iteration mark, we pre-train the $\pi_{\mathrm{distil}}$ policy of RAPID using the method described in Phase 2. Note that $\pi_\mathrm{distil}$ keeps frozen in Phase 3. We report the average performance over $5$ random trails.}
\label{figure:online_adaptation_performance}
\end{figure}

To evaluate the online adaptation ability of RAPID, we employ the pre-trained $\pi_\mathrm{distil}$ from \textbf{Phase~2} (with the collected $\mathcal{D}_\mathrm{off}^\star$ from \textsc{lane-3-density-2}). Then train the $\pi_\mathrm{adapt}$ and its zero gate (as depicted in Fig.~\ref{figure:training-pipeline}(c)) via interacting with different online environments. We compare it with the vanilla Online RL (DQN) and Offline RL (DQN) without using the RAPID MoP policy. We train the RAPID policy for 5K training epochs during the offline phase, followed by 10K online epochs. The Online DQN starts from the 5K-th epoch. 

As depicted in Fig.~\ref{figure:online_adaptation_performance}: (1) The cumulative rewards for $\pi_\mathrm{distil}$ are sufficiently high, due to the introduced common sense knowledge and reasoning abilities from $\pi_\mathrm{LLM}$. However, the conventional offline RL cannot perform well. (2) With the offline phase pre-trained on \textsc{lane-3-density-2}, the online adaptation with RAPID MoP on the same dataset \textsc{lane-3-density-2} (Fig.~\ref{figure:online_adaptation_performance}(a)) achieves significantly high rewards. This demonstrates the necessity of online adaptation to generalize knowledge for practical application. (3) We further conduct zero-shot adaptation on offline-unseen dataset \textsc{lane-4-density-2.5} and \textsc{lane-5-density-3}. In Fig.~\ref{figure:online_adaptation_performance}(b-c), we observe RAPID can achieve competitive performance not only compared to the vanilla online approach, but toward the large $\pi_\mathrm{LLM}$. This highlights the efficacy of the RAPID framework in task adaptation. 

\subsection{Robust Distillation Performance}
\label{section:robust_distillation_performance}

We evaluate the robustness of multiple distillation algorithms using 4 different attack methods, respectively, Uniform, Gaussian, FGSM and PGD. Specifically, we employ a 10-step PGD with a designated step size of $0.01$. For both FGSM and PGD attacks, the attack radius $\epsilon$ is set to $0.1$. For Uniform and Gaussian attacks, $\epsilon$ is set to $0.2$. The observation was normalized before the attack and then denormalized for standard RL policy understanding. In RAPID, we set the $\beta$  as $0.5$ to balance the robustness distillation term according to Eq.~(\ref{eq:robust_distill}). 

As illustrated in Tab.~\ref{table:obs_pert_results}, the conventional methods, Online DQN and Offline DQN, are not able to effectively defend against strong adversarial attacks like FGSM and PGD. Although Offline DQN, which is trained on the dataset $\mathcal{D}_\mathrm{off}^\star$ partially collected by the LLM policy $\pi_{\mathrm{LLM}}$. It still struggles to maintain performance under these attacks. RAPID (w/o $J_{\mathrm{robust}}$), which undergoes offline training and online adaptation without using $J_{\mathrm{robust}}$, performs weakly when facing strong adversarial attacks. In contrast, the full RAPID method with $J_{\mathrm{robust}}$ demonstrates superior robustness against various adversarial attacks across all three environments. We note that the $J_{\mathrm{robust}}$ might approximate the robustness of LLM by building the robust soft label in Eq.~\ref{eq:robust_distill}. Overall, the $J_{\mathrm{robust}}$ regularizer plays a crucial role in bolstering the robustness of the model by effectively utilizing the robust knowledge obtained from the LLM-based teacher. 

\begin{table}[t]
\centering
\caption{Adversarial Performance against observation perturbation, averaged over 5 trails. The Offline DQN, RAPID and RAPID (w/o $J_{\mathrm{robust}}$) are trained based on the dataset $\mathcal{D}_\mathrm{off}^\star$. The RAPID and RAPID (w/o $J_{\mathrm{robust}}$) are processed under offline training and online adaptation, while the Offline DQN only goes through the offline training process. \textbf{Bold} indicates the best return against observation perturbation within each environment.}
\vspace{-5pt}
\resizebox{\linewidth}{!}{\begin{NiceTabular}{@{}c|l|c|cccc@{}}[colortbl-like]
\toprule
\multirow{2}{*}{Environments} 
& \multirow{2}{*}{Method}         
& \multirow{1}{*}{} 
& \multicolumn{4}{c}{Attack Return}                      \\
&  & Clean & Uniform  & Gaussian & FGSM & PGD
\\ \midrule
\multirow{4}{*}{\textsc{lane-3-density-2}}     
& Online DQN & 12.48$_{\pm 8.56}$ 
& 11.74$_{\pm 7.75}$ 
&10.72$_{\pm 8.27}$ 
&2.09$_{\pm 1.62}$ 
&1.09$_{\pm 1.32}$
\\
& Offline DQN      
& 8.66$_{\pm 6.93}$
& 6.73$_{\pm 2.04}$
& 6.29$_{\pm 1.28}$
& 2.12$_{\pm 0.23}$
& 1.73$_{\pm 1.04}$
\\
& \cellcolor{Gray!10}RAPID \tiny{(w/o $J_{\mathrm{robust}}$)} 
& \cellcolor{Gray!10}19.79$_{\pm 3.42}$
& \cellcolor{Gray!10}17.01$_{\pm 4.30}$
& \cellcolor{Gray!10}15.86$_{\pm 5.21}$
& \cellcolor{Gray!10}3.40$_{\pm 1.26}$
& \cellcolor{Gray!10}1.26$_{\pm 1.02}$
\\

& \cellcolor{Gray!30}RAPID  
& \cellcolor{Gray!30}20.42$_{\pm 2.59}$
& \cellcolor{Gray!30}\textbf{20.41$_{\pm 1.58}$}
& \cellcolor{Gray!30}\textbf{20.23$_{\pm 1.26}$}
& \cellcolor{Gray!30}\textbf{15.34$_{\pm 2.97}$}
& \cellcolor{Gray!30}\textbf{14.77$_{\pm 1.23}$}

\\
\midrule
\multirow{4}{*}{\textsc{lane-4-density-2.5}}   
& Online DQN & 11.58$_{\pm 0.56}$
& 10.49$_{\pm 4.52}$
& 10.18$_{\pm 4.72}$
& 6.03$_{\pm 0.86}$
& 6.12$_{\pm 0.78}$
\\
& Offline DQN 
& 7.14$_{\pm 6.12}$
& 5.20$_{\pm 1.47}$
& 6.56$_{\pm 1.06}$
& 2.56$_{\pm 1.06}$
& 2.20$_{\pm 1.47}$

\\
& \cellcolor{Gray!10}RAPID \tiny{(w/o $J_{\mathrm{robust}}$)} 
& \cellcolor{Gray!10}13.79$_{\pm 3.48}$
& \cellcolor{Gray!10}12.87$_{\pm 1.43}$
& \cellcolor{Gray!10}11.61$_{\pm 5.63}$
& \cellcolor{Gray!10}5.53$_{\pm 3.27}$
& \cellcolor{Gray!10}3.26$_{\pm 1.24}$

\\
& \cellcolor{Gray!30}RAPID 
& \cellcolor{Gray!30}14.34$_{\pm 2.84}$
& \cellcolor{Gray!30}\textbf{13.63$_{\pm 5.26}$}
& \cellcolor{Gray!30}\textbf{13.91$_{\pm 7.84}$}
& \cellcolor{Gray!30}\textbf{10.87$_{\pm 2.68}$}
& \cellcolor{Gray!30}\textbf{8.39$_{\pm 2.11}$}

\\
                            \midrule
\multirow{4}{*}{\textsc{lane-5-density-3}}      
& Online DQN 
& 6.05$_{\pm 9.41}$
& \textbf{6.64$_{\pm 1.46}$}
& \textbf{5.48$_{\pm 7.67}$}
& 1.55$_{\pm 1.10}$
& 1.54$_{\pm 1.11}$
\\

& Offline DQN   
& 5.99$_{\pm 4.10}$ 
& 2.55$_{\pm 2.10}$
& 2.22$_{\pm 1.99}$
& 1.22$_{\pm 1.03}$
& 0.88$_{\pm 0.58}$

\\

& \cellcolor{Gray!10}RAPID \tiny{(w/o $J_{\mathrm{robust}}$)} 
& \cellcolor{Gray!10}8.47$_{\pm 2.64}$
& \cellcolor{Gray!10}4.28$_{\pm 1.44}$
& \cellcolor{Gray!10}3.20$_{\pm 2.06}$
& \cellcolor{Gray!10}0.34$_{\pm 0.24}$
& \cellcolor{Gray!10}0.79$_{\pm 0.33}$
 
\\

& \cellcolor{Gray!30}RAPID 
& \cellcolor{Gray!30}7.83$_{\pm 3.41}$
& \cellcolor{Gray!30}5.14$_{\pm 3.19}$
& \cellcolor{Gray!30}5.22$_{\pm 2.68}$
& \cellcolor{Gray!30}\textbf{2.96$_{\pm 0.76}$}
& \cellcolor{Gray!30}\textbf{1.62$_{\pm 0.48}$}

\\

\bottomrule
\end{NiceTabular}}
\label{table:obs_pert_results}
\end{table}

\subsection{Ablation Study}

\begin{figure}[tb]
\centering
\includegraphics[width=1\linewidth]{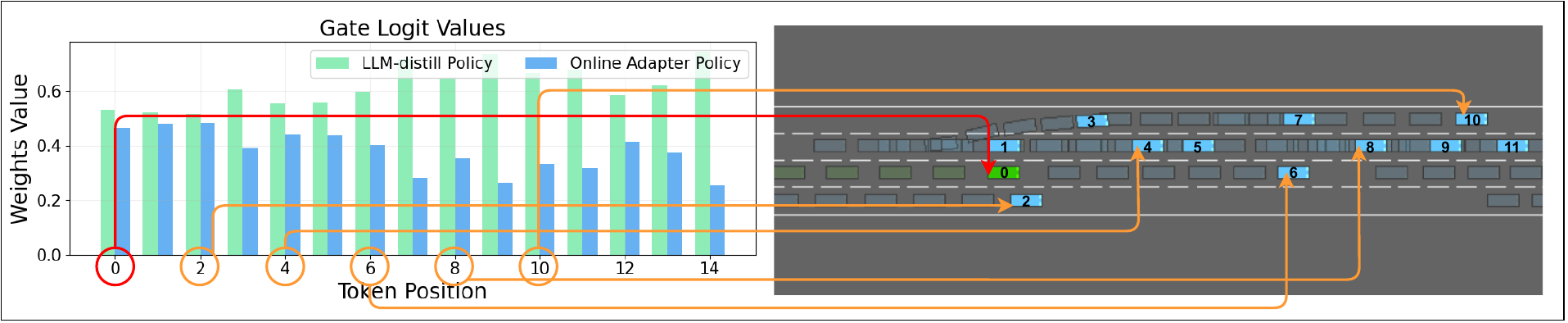}
\vfill
\caption{Visualizing the contribution of $\pi_\mathrm{distil}$ and $\pi_\mathrm{adapt}$ to the final predicted action. The example is randomly sampled via interacting with \textsc{lane-4-density-2.5} using $\Pi_\mathrm{MoP}$. More samples are demonstrated in Appendix.~\ref{app:visualization_policies}.}
\label{figure:gate_weights}
\end{figure}
\textbf{MoP Routing Analysis. }In Fig.~\ref{figure:gate_weights}, we visualize the contribution of each policy ($\pi_\mathrm{distil}$ and $\pi_\mathrm{adapt}$) to the final predicted action at each vehicle position after online adaptation. In general, $\pi_\mathrm{distil}$ dominates the contribution across all vehicles, as $\pi_\mathrm{adapt}$ is initially masked out during offline training. However, after the online phase, $\pi_\mathrm{adapt}$ provides a more balanced contribution, especially for the ego vehicle and nearby vehicles (like vehicles 0, 2, 4 in Fig.~\ref{figure:gate_weights}), indicating the effectiveness of online adaptation. Interestingly, $\pi_\mathrm{distil}$ still contributes substantially for distant vehicles, suggesting the value of LLM-distilled knowledge in understanding overall traffic context. This demonstrates the efficacy of the MoP approach in integrating knowledge from both the LLM-based teacher and the environment interaction, leveraging the most relevant expertise for each vehicle based on the domain and relative position.

\textbf{Additional Ablation Studies.} Please refer to Appendix~{\ref{app:additional_ablation}}.





\section{Conclusion}
\vspace{-5pt}

We propose RAPID, a promising approach for leveraging LLMs' reasoning abilities and common sense knowledge to enhance the performance of RL agents in heterogeneous autonomous driving tasks. Meanwhile, RAPID overcomes challenges such as the long inference time of LLMs and policy knowledge overwriting. The robust knowledge distillation method enables the student RL policy to inherit the robustness of the LLM teacher. 

\textbf{Limitation and future work.} Although we have conducted tests in three distinct autonomous driving environments and validated the closed-loop RL policy in real-time, the scope of the analysis is restricted. The 2D HighwayEnv remains overly simplistic for comprehensive autonomous driving evaluation. To further establish the efficacy of our online adaptation policy, it is necessary to assess its performance with Visual Language Models \cite{sima2023drivelm,wang2023drivemlm} in more realistic and complex autonomous driving environments, like CARLA \cite{dosovitskiy2017carla}. 


\clearpage
\acknowledgments{If a paper is accepted, the final camera-ready version will (and probably should) include acknowledgments. All acknowledgments go at the end of the paper, including thanks to reviewers who gave useful comments, to colleagues who contributed to the ideas, and to funding agencies and corporate sponsors that provided financial support.}


\bibliography{reference}  

\clearpage

\appendix
\onecolumn
\section{Related Work}
\subsection{Offline RL for Autonomous Driving}
Offline RL algorithms are designed to learn policies from a static dataset, eliminating the need for interaction with the real environment \cite{levine2020offline}. Compared to conventional online RL \cite{mnih2013playing} and the extended goal-conditioned RL \cite{yin2023rerogcrl}, this approach is especially beneficial in scenarios where such interaction is prohibitively expensive or risky, such as in autonomous driving. Offline RL algorithms have demonstrated the capability to surpass expert-level performance \cite{fujimoto2019offpolicy,kumar2020conservative,kostrikov2022offline}. In general, these algorithms employ policy regularization \cite{fujimoto2019offpolicy,fujimoto2021minimalist} and out-of-distribution (OOD) penalization \cite{kumar2020conservative,wu2021uncertainty} as strategies to prevent value overestimation. In this paper, we pioneer utilising the LLM-generated data to train the offline RL. Although there have been several works \cite{gu2023knowledge,hsieh2023distilling} focus on distillation for LLM, none of them considers distilling to RL. 


\subsection{LLM for Autonomous Driving}
Recent advancements in LLMs \cite{brown2020language,GPT4,touvron2023llama} demonstrate their powerful embodied abilities, providing the possibility to distil knowledge from humans to autonomous systems. LLMs exhibit a strong aptitude for general reasoning \cite{zhou2023language,lin2023swiftsage}, web agents \cite{deng2023mind2web,zhou2023webarena}, and embodied robotics \cite{nottingham2023embodied,song2023llm,wang2023describe}. Inspired by the superior capability of common sense of LLM-based agents, a substantial body of research is dedicated to LLM-based autonomous driving.  Wayve \cite{Lingo-1} introduced an open-loop driving commentator called LINGO-1, which integrates vision, language, and action to enhance the interpretation and training of driving models. DiLu \cite{wen2023dilu} developed a framework utilizing LLMs as agents for closed-loop driving tasks, with a memory module to record experiences. To enhance the stability and generalization performance, \cite{fu2023drive}
utilised reasoning, interpretation, and memorization of LLM to enhance autonomous driving. More research works for this vibrant field were summarised in \cite{yang2023survey}. However, all these methods require huge resources, long inference time, and unstable performance. To bridge this gap, we propose a knowledge distillation framework, from LLM to RL, which enhances the applicability and stability of real-world autonomous driving. 

\subsection{Distillation for LLM}
Knowledge distillation has proven successful in transferring knowledge from large-scale, more competent teacher models to small-scale student models, making them more affordable for practical applications
\cite{pmlr-v202-fu23d,beyer2022knowledge,west-etal-2022-symbolic}. This method facilitates learning from limited labelled data, as the larger teacher model is commonly employed to generate a training dataset with noisy pseudo-labels \cite{iliopoulos2022weighted,smith2022language,wang-etal-2021-want-reduce}.
Further, \cite{hsieh2023distilling} involves extracting rationales from LLMs as additional supervisory signals to train small-scale models within a multi-task framework. \cite{gu2023knowledge} utilises reverse Kullback-Leibler divergence to ensure that the student model does not overestimate the low-probability regions in the teacher distribution. Currently, knowledge distillation from LLM to RL for autonomous driving remains unexplored.

\newpage
\section{Additional Experiments}
\label{app:additional_ablation}

\subsection{Comparison between MLP and RAPID(Attentive) based Policy}

\begin{figure}[H]
\centering
\includegraphics[width=1\linewidth]{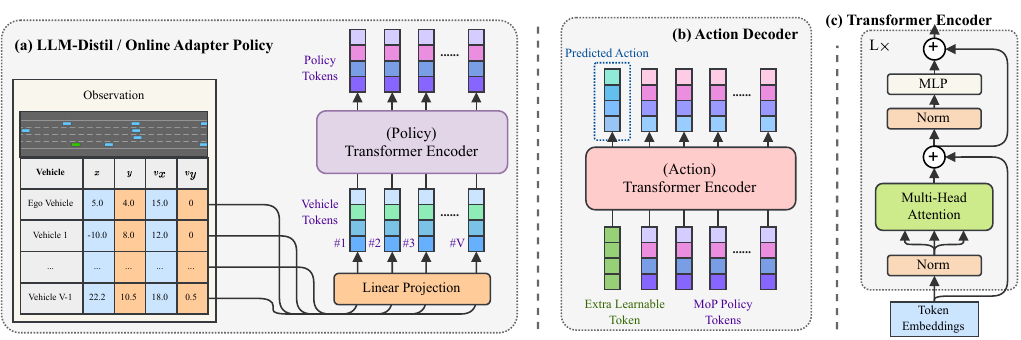}
\caption{Architecture of (a) Policy networks (transformer encoders) $\pi_\mathrm{distil}(\cdot)$ and $\pi_\mathrm{adapt}(\cdot)$; (b) Action decoder (transformer encoder) $\mathrm{dec}(\cdot)$; (c) The transformer backbone (encoder-only).}
\label{fig:modules_architectures}
\end{figure}

\begin{table}[H]
\centering
\caption{Comparison between MLP and RAPID(Attentive) based policy under offline RL training. \textbf{Bold} denotes the best among each environment. Results are averaged across 3 seeds. 
}
\resizebox{\linewidth}{!}
{\begin{NiceTabular}{@{}c|c|ccc|ccc@{}}[colortbl-like]
\toprule
\multirow{2}{*}{Environments}                                 & \multirow{2}{*}{Dataset}  & \multicolumn{3}{c|}{MLP Policy} & \multicolumn{3}{c}{RAPID(Attentive) Policy} 
\\ & & CQL                
& DQN                
& DDQN               
& CQL 
& DQN 
& DDQN                                    \\ \midrule
\multirow{3}{*}{\textsc{lane-3-density-2}}   
& $\mathcal{D}_\mathrm{rand}$                
& 3.01$_{\pm 1.49}$  
&  4.40$_{\pm 1.57}$  
& 4.86$_{\pm 1.58}$  
& 3.50$_{\pm 0.84}$                        
& 3.66$_{\pm 1.58}$         
& 2.56$_{\pm 0.18}$ 
\\
& $\mathcal{D}_\mathrm{LLM}$                 
& 7.96$_{\pm 1.63}$  
& 9.14$_{\pm 8.80}$  
& 10.83$_{\pm 4.94}$ 
& 12.81$_{\pm 7.51}$      
& 14.18$_{\pm 2.43}$     
& 5.65$_{\pm 3.13}$
\\
& $\mathcal{D}_\mathrm{off}^\star$           
& 13.06$_{\pm 1.11}$ 
& 8.66$_{\pm 6.93}$  
& 12.22$_{\pm 7.03}$ 
& 12.84$_{\pm 0.68}$                       
& \textbf{15.33$_{\pm 5.30}$} 
& 13.95$_{\pm 1.27}$
\\ \midrule
\multirow{3}{*}{\textsc{lane-4-density-2.5}} 
& $\mathcal{D}_\mathrm{rand}$                   
& 3.84$_{\pm 2.59}$  
& 2.41$_{\pm 0.92}$  
& 1.96$_{\pm 0.52}$  
& 6.69$_{\pm 3.82}$                
& 3.60$_{\pm 2.67}$
& 1.70$_{\pm 0.46}$
\\                                             
& $\mathcal{D}_\mathrm{LLM}$                    
& 2.56$_{\pm 0.90}$  
& 3.99 $_{\pm 1.18}$ 
& 5.82$_{\pm 6.16}$  
& 5.98$_{\pm 1.30}$                
& 4.60$_{\pm 2.69}$ 
& 2.23$_{\pm 0.72}$ \\
& $\mathcal{D}_\mathrm{off}^\star$    
& 7.36$_{\pm 0.60}$  
& 7.14$_{\pm 6.12}$  
& 7.61$_{\pm 6.76}$  
& 8.14$_{\pm 0.87}$             
& \textbf{10.29$_{\pm 0.20}$} 
& 3.16$_{\pm 0.31}$
\\ \midrule
\multirow{3}{*}{\textsc{lane-5-density-3}}   
& $\mathcal{D}_\mathrm{rand}$                   
& 1.53$_{\pm 0.29}$  
& 5.59$_{\pm 2.84}$  
& 3.72$_{\pm 2.85}$  
& 2.58$_{\pm 1.09}$               
& 2.16$_{\pm 0.36}$   
& 1.83$_{\pm 1.48}$   \\
& $\mathcal{D}_\mathrm{LLM}$                    
& 2.23$_{\pm 0.29}$  
& 5.54$_{\pm 7.98}$  
& 4.76$_{\pm 1.89}$  
& 2.28$_{\pm 1.65}$          
& 1.45$_{\pm 2.11}$    
& 2.33$_{\pm 0.66}$    
\\
& $\mathcal{D}_\mathrm{off}^\star$                        
& 5.58$_{\pm 3.93}$  
& 5.99$_{\pm 4.10}$  
& 5.05$_{\pm 2.06}$  
& 3.37$_{\pm 1.61}$                
& \textbf{6.14$_{\pm 2.85}$}    
& 3.17$_{\pm 0.41}$
\\ \bottomrule
\end{NiceTabular}}
\label{table:mlp_transformer_com}
\end{table}
To verify the contribution of the RAPID(Attentive) architecture (as shown in Fig.~\ref{fig:modules_architectures}) in the offline training phase, we conduct extra experiments in this section. As illustrated in Tab.~\ref{table:mlp_transformer_com}, we compared the DQN, DDQN, and CQL under MLP and RAPID architecture, respectively. As the environment complexity increases, the performance gap between RAPID and MLP narrows, suggesting RAPID handles simpler environments more effectively. In summary, we conlude: (1) the RAPID + DQN method achieves the best performance among all methods, thus we choose DQN as the backbone of RAPID for offline training. (2) The attentive architecture demonstrates superior performance compared to MLP, particularly in less complex environments.

\subsection{Impact of $J_{\mathrm{robust}}$}

\begin{figure}[H]
\centering
\includegraphics[width=1\linewidth]{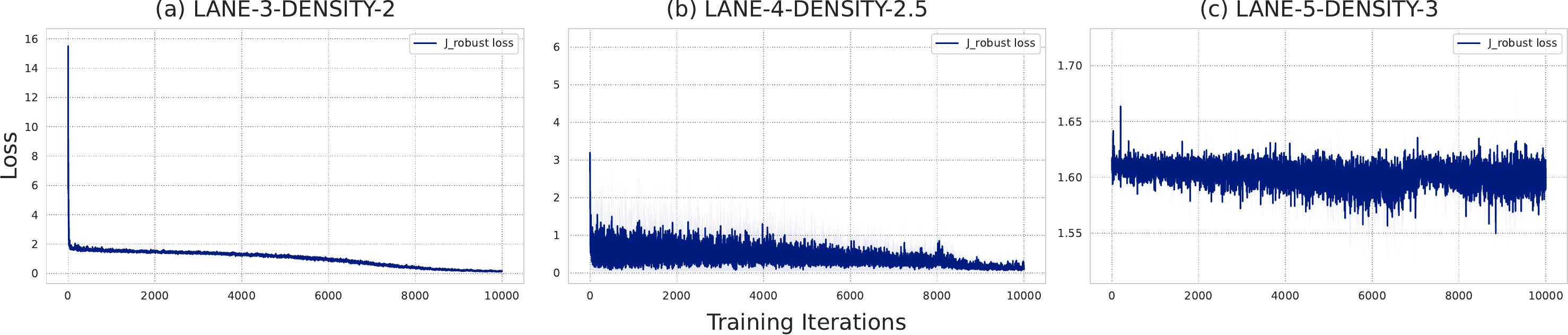}
\caption{Offline training $J_{\mathrm{robust}}$ loss term under the \textsc{lane-3-density-2}-$\mathcal{D}_\mathrm{off}^\star$.}
\label{fig:j_loss_app}
\end{figure}
Building upon the results presented in Fig.~\ref{fig:j_loss_app}, we provide an analysis of the impact of $J_{\mathrm{robust}}$ among three environments. The experiment setup is the same as Sec.~\ref{section:distillation_benchmark}. The RAPID is trained 10K epochs based on \textsc{lane-3-density-2}-$\mathcal{D}_\mathrm{off}^\star$, \textsc{lane-4-density-2.5}-$\mathcal{D}_\mathrm{off}^\star$, and \textsc{lane-5-density-3}-$\mathcal{D}_\mathrm{off}^\star$. From Fig.~\ref{fig:j_loss_app}, we observe that the $J_{\mathrm{robust}}$ gradually decreased and converged to 0 during the training process. The experimental results align with our results in Tab.~\ref{table:obs_pert_results}, as the defense mechanism demonstrates superior performance in both \textsc{lane-3-density-2} and \textsc{lane-4-density-2.5}. In the more complex \textsc{lane-5-density-3} environment, the loss remains relatively high, around 1.6, which leads to suboptimal defense performance compared to the other scenarios. This suggests that the increased complexity and vehicle density in this setting pose additional challenges for the defense mechanism. Overall, the consistency across different settings highlights the robustness and effectiveness of our approach. 

\vfill

\subsection{Impact of $\beta$}
\begin{table}[H]
\centering
\caption{Adversarial Performance of different $\beta$ parameters against observation perturbation.}
\label{table:beta_impact}
\resizebox{\linewidth}{!}
{\begin{NiceTabular}{@{}c|l|c|cccc@{}}[colortbl-like]
\toprule
\multirow{2}{*}{Environments} 
& \multirow{2}{*}{Method}         
& \multirow{1}{*}{} 
& \multicolumn{4}{c}{Attack Return}                      \\
&  & Clean & Uniform  & Gaussian & FGSM & PGD
\\ \midrule
\multirow{4}{*}{\textsc{lane-3-density-2}}     

& RAPID \tiny{(w/o $J_{\mathrm{robust}}$)} 
& 19.79$_{\pm 3.42}$
& 17.01$_{\pm 4.30}$
& 15.86$_{\pm 5.21}$
& 3.40$_{\pm 1.26}$
& 1.26$_{\pm 1.02}$
\\

& RAPID ($\beta = 0.1$)
& 20.74$_{\pm 3.66}$
& 21.22$_{\pm 2.95}$
& 20.75$_{\pm 3.43}$
& 10.97$_{\pm 3.23}$
& 10.24$_{\pm 2.67}$

\\
& RAPID ($\beta = 0.5$)
& 20.42$_{\pm 2.59}$
& 20.41$_{\pm 1.58}$
& 20.23$_{\pm 1.26}$
& 15.34$_{\pm 2.97}$
& 14.77$_{\pm 1.23}$

\\
& RAPID ($\beta = 0.8$) 
& 18.36$_{\pm 3.46}$
& 18.23$_{\pm 3.98}$
& 17.68$_{\pm 2.54}$
& 13.51$_{\pm 4.14}$
& 11.68$_{\pm 4.11}$

\\
\midrule
\multirow{4}{*}{\textsc{lane-4-density-2.5}}   

& RAPID \tiny{(w/o $J_{\mathrm{robust}}$)} 
& 13.79$_{\pm 3.48}$
& 12.87$_{\pm 1.43}$
& 11.61$_{\pm 5.63}$
& 5.53$_{\pm 3.27}$
& 3.26$_{\pm 1.24}$

\\
& RAPID ($\beta = 0.1$)
& 14.12$_{\pm 1.76}$
& 12.66$_{\pm 4.87}$
& 11.38$_{\pm 6.29}$
& 8.17$_{\pm 3.75}$
& 7.13$_{\pm 2.42}$
\\
& RAPID ($\beta = 0.5$)
& 14.34$_{\pm 2.84}$
& 13.63$_{\pm 5.26}$
& 13.91$_{\pm 7.84}$
& 10.87$_{\pm 2.68}$
& 8.39$_{\pm 2.11}$
\\
& RAPID ($\beta = 0.8$)
& 14.21$_{\pm 5.23}$
& 12.24$_{\pm 3.97}$
& 12.68$_{\pm 4.25}$
& 8.69$_{\pm 3.24}$
& 9.16$_{\pm 4.13}$
\\
                            \midrule
\multirow{4}{*}{\textsc{lane-5-density-3}}      



& RAPID \tiny{(w/o $J_{\mathrm{robust}}$)} 
& 8.47$_{\pm 2.64}$
& 4.28$_{\pm 1.44}$
& 3.20$_{\pm 2.06}$
& 0.34$_{\pm 0.24}$
& 0.79$_{\pm 0.33}$
 
\\

& RAPID ($\beta = 0.1$)
& 6.98$_{\pm 4.84}$
& 6.29$_{\pm 4.59}$
& 6.21$_{\pm 2.96}$
& 1.90$_{\pm 1.67}$
& 2.19$_{\pm 1.23}$

\\
& RAPID ($\beta = 0.5$)
& 7.83$_{\pm 3.41}$
& 5.14$_{\pm 3.19}$
& 5.22$_{\pm 2.68}$
& 2.96$_{\pm 0.76}$
& 1.62$_{\pm 0.48}$

\\
& RAPID ($\beta = 0.8$)
& 6.17$_{\pm 2.19}$
& 4.13$_{\pm 4.28}$
& 5.77$_{\pm 3.56}$
& 2.89$_{\pm 0.32}$
& 2.67$_{\pm 0.89}$

\\

\bottomrule
\end{NiceTabular}}
\end{table}
To investigate the effect of various $\beta$ values (hyperparameter in Eq.~\ref{eq:robust_distill}), we compare $\beta\in\{0.1, 0.5, 0.8\}$ on \textsc{lane-3-density-2}, \textsc{lane-4-density-2.5}, and \textsc{lane-5-density-3}, respectively. The results are illustrated in Tab.~\ref{table:beta_impact}. Based on the results presented in Tab.~\ref{table:beta_impact}, we make several observations regarding the impact of the $\beta$ on the performance of our robust distillation approach. When setting $\beta=0.8$, we notice a decline in the clean performance across all three environments. This degradation can be attributed to the increased emphasis on adversarial examples during the training process, which may lead to a trade-off with clean accuracy. On the other hand, when setting $\beta=0.1$, the attack return is relatively lower compared to other settings. This suggests that the adversarial training strength may not be sufficient to provide adequate robustness against adversarial perturbations. Considering these findings, we determine that setting $\beta=0.5$ strikes a balance between maintaining clean performance and achieving satisfactory robustness. 

\vfill

\subsection{Visualization the of $\pi_\mathrm{distil}$ and $\pi_\mathrm{adapt}$}\label{app:visualization_policies}

\begin{figure}[H]
\centering
\includegraphics[width=1\linewidth]{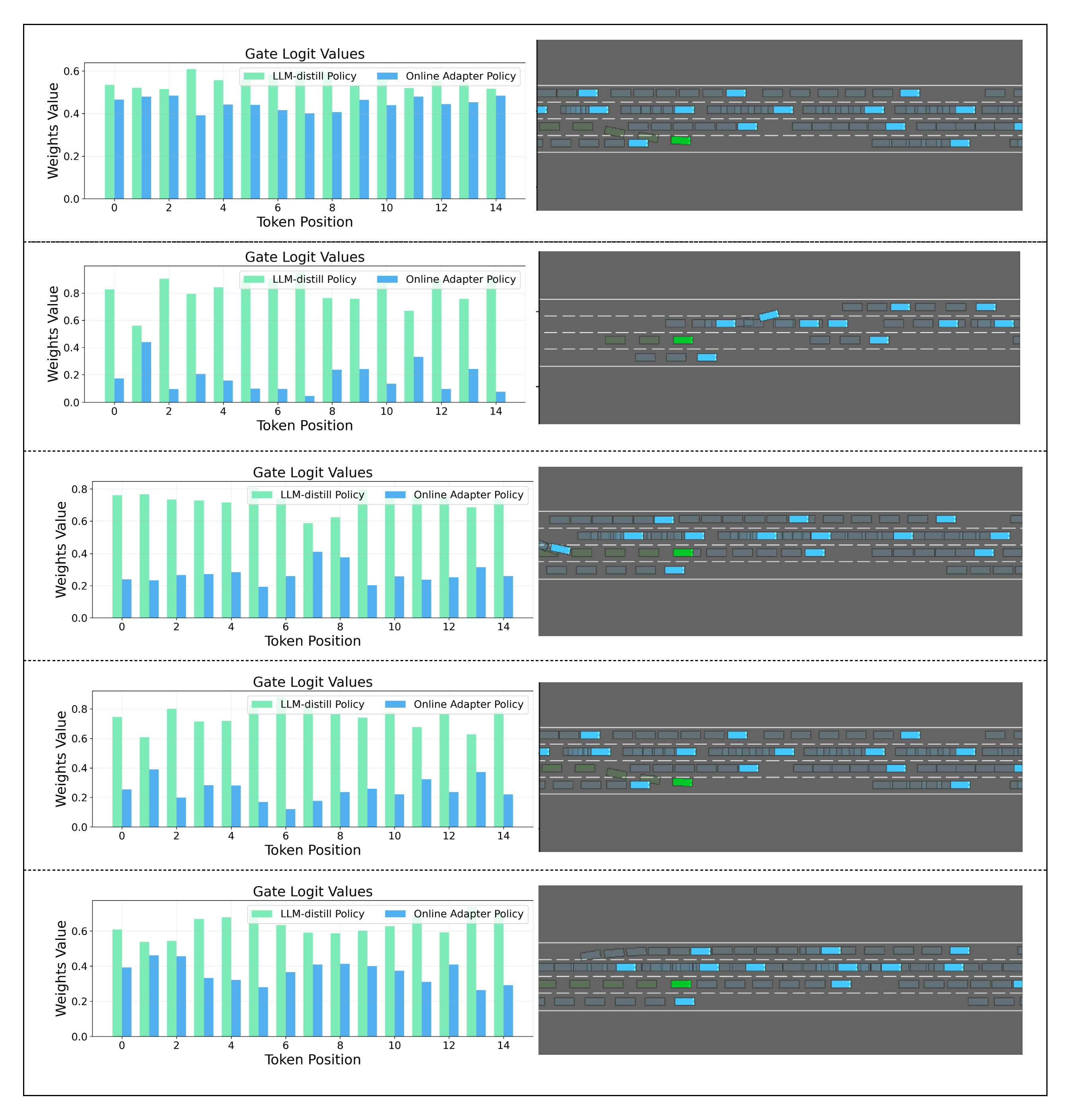}
\caption{Five more random examples on visualizing the contribution of $\pi_\mathrm{distil}$ and $\pi_\mathrm{adapt}$ (from the router) to the final predicted action in \textsc{lane-4-density-2.5} using $\Pi_\mathrm{MoP}$.}
\end{figure}

\clearpage
\section{Network Implementation}

\label{app:nets_implementation}

\subsection{Transformer Encoder}
We employ the same encoder-only transformer $f_\mathrm{TransE}$ as \cite{vaswani2017attention}.

\subsection{Policy Networks}

The policy network consist of a linear projection and a transformer encoder that take the projected state as input. Let state $\mathbf{s}\in \mathbb{R}^{V\times F}$. The policy network first project $\mathbf{s}\to\mathbf{s}_v\in \mathbb{R}^{V\times {F'}}$ with a trainable linear projection, then regard $\mathbf{s}_v$ as the input token for the transformer. The transformer then processes these embeddings through $[0,L]$ layers of self-attention and feedforward networks, in our case, we set $L=2$. Mathematically, for each layer \( l\in[0,L] \), the $\pi_\mathrm{distil}/\pi_\mathrm{adapt}$ computes the following: $\mathbf{s}_p \gets f_{\mathrm{TransE}}^\mathrm{policy}(f_{\mathrm{Proj}}(\mathbf{s})) \in \mathbb{R}^{V\times D}$.

\subsection{Action Decoder}

The action decoder transforms encoded state representations into an action using another transformer encoder. It takes the mix-of-policy token from the routed policy networks as input. Given the routed MoP token $\mathbf{s}_{m} \in \mathbb{R}^{V\times {D}}$, we first concatenate $\mathbf{s}_{m}$ with an extra learnable token $\mathbf{s}_e\in \mathbb{R}^{1\times {D}}$, then put $\mathrm{concat}(\mathbf{s}_m\| \mathbf{s}_e)\in\mathbb{R}^{(V+1)\times {D}}$ into the transformer encoder. We regard the first output token as the action token. Essentially, $\mathrm{dec}:\mathbb{R}^{V\times D}\to \mathbb{R}^A$ computes: $\mathbf{a} \gets f_{\mathrm{TransE}}^\mathrm{action}(\mathrm{concat}(\mathbf{s}_m\| \mathbf{s}_e))_{0,:} \in \mathbb{R}^{1\times {A}}$.

\section{Environment Details}
\label{app:env_details}

\begin{figure}[H]
\centering
\includegraphics[width=0.4\linewidth]{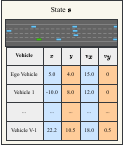}
\caption{State features under KinematicObservation setting.}
\label{figure:state_feature}
\end{figure}

Each of these three environments has continuous state and discrete action space. The maximum episode horizon is defined as 30. Tab.~\ref{tab:highway_config} describes the configuration for the highway environment in a reinforcement learning setting. The observation configuration defines the type of observation utilized, which is specified as KinematicsObservation. This observation type represents the surrounding vehicles as a $V \times F$ array, where $V$ denotes the number of nearby vehicles and $F$ represents the size of the feature set describing each vehicle. The specific features included in the observation are listed in the features field of the configuration.
The KinematicsObservation provides essential information about the neighbouring vehicles, such as their presence, positions in the x-y coordinate system, and velocities along the x and y axes. These features are represented as absolute values, independent of the agent's frame of reference, and are not subjected to any normalization process. Fig.~\ref{figure:state_feature} demonstrates the state features under KinematicObservation setting, in which we introduce the vehicle feature tokenization in Sec.~\ref{RAPID}. 


\begin{table}[htb]
\centering
\caption{Highway Environment Configuration}
\label{tab:highway_config}
\begin{tabular}{l|l}
\toprule
\textbf{Parameter} & \textbf{Value} \\ \midrule
Observation Type & KinematicsObservation \\
Observation Features & presence, $x$, $y$, $v_x$, $v_y$ \\
Observation Absolute & True \\
Observation Normalize & False \\
Observation Vehicles Count & 15 \\
Observation See Behind & True \\
Action Type & DiscreteMetaAction \\
Action Target Speeds & np.linspace(0, 32, 9) \\
Lanes Count & 3, (4, 5) \\
Duration & 30 \\
Vehicles Density & 2, (2.5, 3) \\
Show Trajectories & True \\
Render Agent & True \\
\bottomrule
\end{tabular}
\end{table}

\section{Hyperparameters}
\label{app:hyperparameters}

In this work, conventional DQN, DDQN and CQL share the same $\mathrm{MLP}$ network
architectures.  For all experiments, the hyperparameters of our backbone architectures and algorithms are reported in Tab. ~\ref{table:hyperparameter}. Our implementation is based on d3rlpy \cite{seno2022d3rlpy}, which is open-sourced.

\begin{table}[h]
\centering
\caption{Hyperparameters for conventional Offline RL methods.}
\label{table:hyperparameter}
\resizebox{\linewidth}{!}{\begin{tabular}{@{}c|c|c|c|c|c|c|c|c@{}}
\toprule
& \textbf{Hyperparameter} & \textbf{Value} & & \textbf{Hyperparameter} & \textbf{Value} & & \textbf{Hyperparameter} & \textbf{Value}     \\ \midrule
\multirow{1}{*}{DQN} & Batch size               & 32  & \multirow{1}{*}{DDQN}  & Batch size & 32 & \multirow{1}{*}{CQL}  & Batch size & 64         \\
& Learning rate     & \(5 \times 10^{-3}\) & & Learning rate & \(5 \times 10^{-4}\) &
& Learning rate & \(5 \times 10^{-4}\) \\
& Target network update interval & 50 
& 
& Target network update interval & 50 &
& Target network update interval & 50 \\
& Total timesteps        & 10000  &
& Total timesteps & 10000 &
& Total timesteps & 10000\\
& Timestep per epoch        & 100 &
& Timestep per epoch & 100 & 
& Timestep per epoch & 100 \\
& Buffer capacity             & 1$\times10^5$   &
& Buffer capacity             & 1$\times10^5$  &
& Buffer capacity             & 1$\times10^5$  
\\
& Number of Tests & 10 & 
& Number of Tests & 10 &
& Number of Tests & 10\\
& Critic hidden dim & 256 &
& Critic hidden dim & 256 &
& Critic hidden dim & 256\\
& Critic hidden layers & 2 & 
& Critic hidden layers & 2 &
& Critic hidden layers & 2 \\
& Critic activation function & ReLU &
& Critic activation function & ReLU &
& Critic activation function & ReLU\\
\bottomrule
\end{tabular}

}
\end{table}

\section{Prompt Setup}\label{app:prompt_setup}




In this section, we detail the specific prompt design and give an example of the interaction between the LLM-based agent and the environment. 

\begin{figure}[ht]
\centering
\includegraphics[width=1\linewidth]{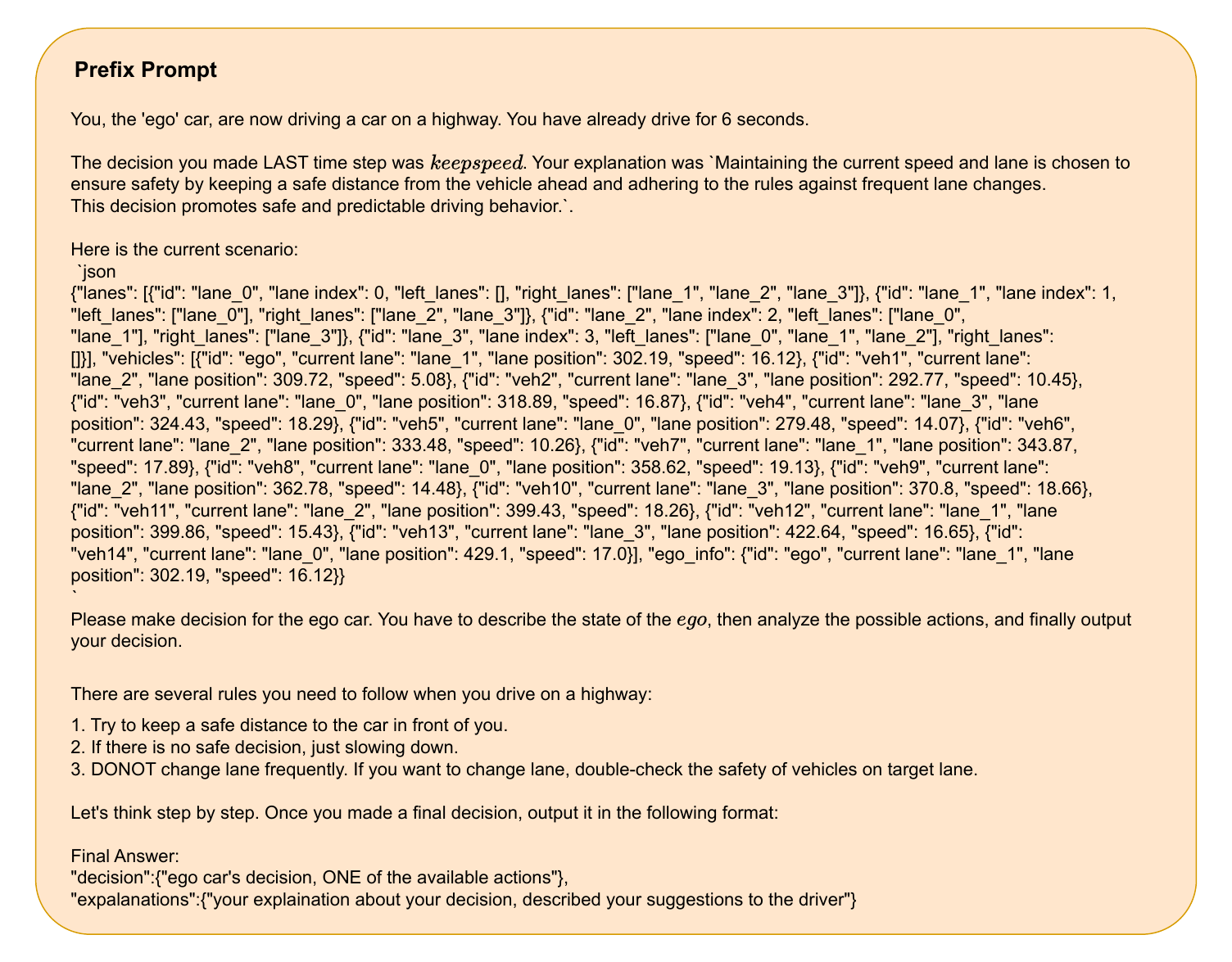}
\caption{Prefix Prompt before interacting with HighwayEnv.}
\label{figure:prefix_prompt}
\end{figure}

\textbf{Prefix Prompt.} As shown in Fig.~\ref{figure:prefix_prompt}, the Prefix Prompt part primarily consists of an introduction to the autonomous driving task, a description of the scenario, common sense rules, and instructions for the output format. The previous decision and explanation are obtained from the experience buffer. The current scenario information plays an important role while making decision, and it is dynamically generated based on the current decision frame.  The driving
scenario description contains information about the ego and surrounding vehicles' speed and positions. The common sense rules section embeds the driving style to guide the vehicle's behaviour. Finally, the final answer format is constructed to output the discrete actions and construct the closed-loop simulation on HighwayEnv.

\begin{figure}[ht]
\centering
\includegraphics[width=1\linewidth]{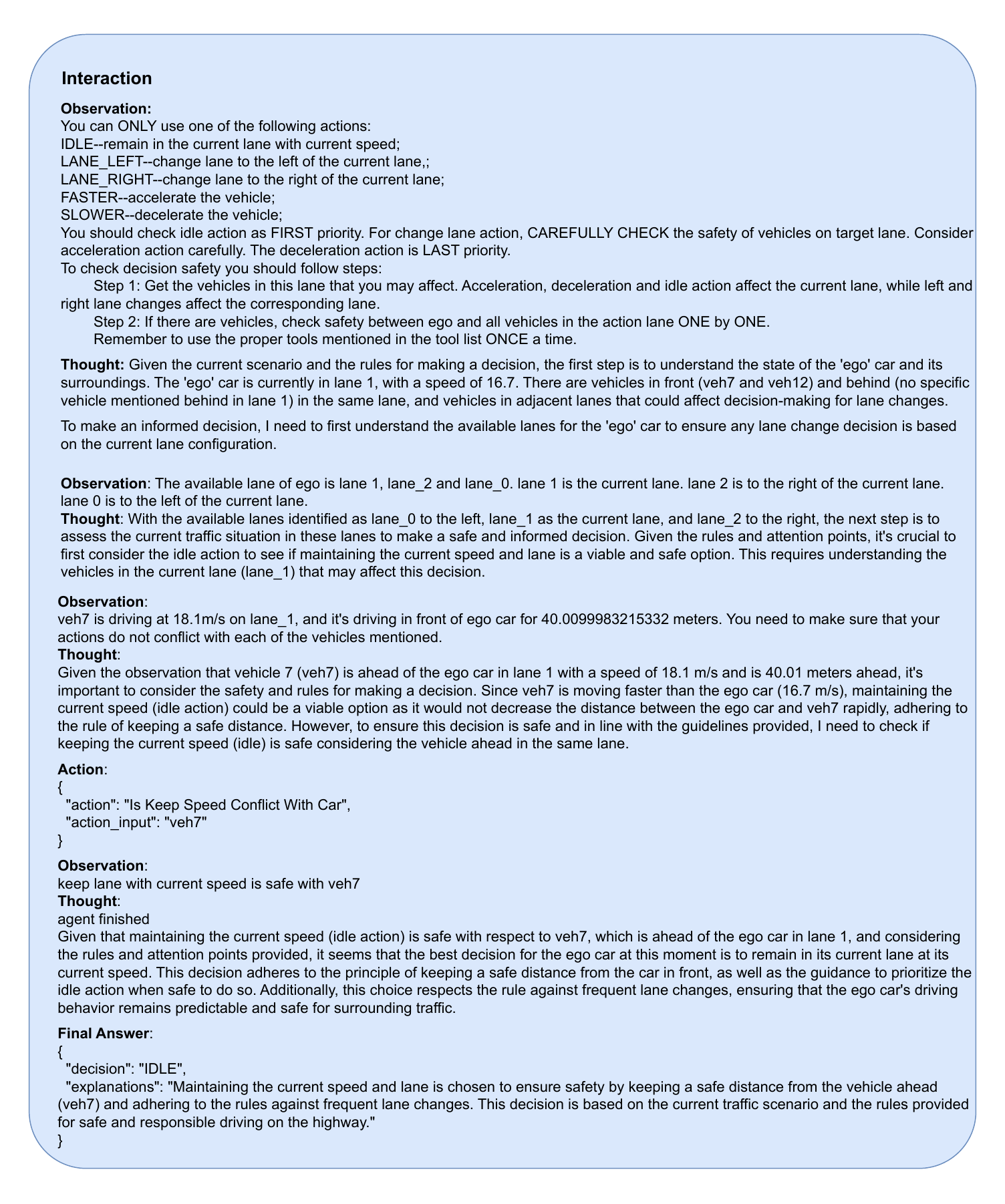}
\caption{Interaction between the LLM-based agent and the HighwayEnv environment.}
\label{figure:prompt_interaction}
\end{figure}

\textbf{Interaction.} We demonstrate one example to make readers better understand the reasoning process of \texttt{GPT-3.5}. As shown in Fig.~\ref{figure:prompt_interaction}, the ego car initially checks the available actions and related safety outcomes. On the first round of thought, \texttt{GPT-3.5} tries to understand the situation of the ego car and checks the available lanes for decision-making. After several rounds of interaction, it checks whether the action \texttt{keep speed} is safe with vehicle 7. Finally, it outputs the decision \texttt{idle} and explains that maintaining the current speed and lane can keep a safe distance from surrounding cars.   

\end{document}